# Vibration-Based Condition Monitoring By Ensemble Deep Learning

Vahid YAGHOUBI [1,2], Liangliang CHENG [1,2], Wim VAN PAEPEGEM [1], Mathias KERSEMANS [1]
[1]Mechanics of Materials and Structures (MMS), Ghent University,
Technologiepark 46, B-9052 Zwijnaarde, Belgium.
e-mail: vahid.yaghoubi@ugent.be, Liangliang.cheng@ugent.be, wim.vanpaepegem@ugent.be, mathias.kersemans@ugent.be
[2]SIM M3 program, Technologiepark 48 , B-9052 Zwijnaarde, Belgium.

**Abstract**
Vibration-based techniques are among the most common condition monitoring approaches. With the advancement of computers, these approaches have also been improved such that recently, these approaches in conjunction with deep learning methods attract attention among researchers. This is mostly due to the nature of the deep learning method that could facilitate the monitoring procedure by integrating the feature extraction, feature selection, and classification steps into one automated step. However, this can be achieved at the expense of challenges in designing the architecture of a deep learner, tuning its hyper-parameters. Moreover, it sometimes gives low generalization capability. As a remedy to these problems, this study proposes a framework based on ensemble deep learning methodology. The framework was initiated by creating a pool of Convolutional neural networks (CNN). To create diversity to the CNNs, they are fed by frequency responses which are passed through different functions. As the next step, proper CNNs are selected based on an information criterion to be used for fusion. The fusion is then carried out by improved Dempster-Shafer theory. The proposed framework is applied to real test data collected from Equiax Polycrystalline Nickel alloy first-stage turbine blades with complex geometry.

## 1. Introduction

Vibration-based is one of the most common condition monitoring techniques that has been evolved from traditional methods to machine learning (ML)-based methods [1]. Deep learning (DL) is a new trend among researchers in the domain of condition monitoring due to its promising result and automated feature learning [2,3]. An extensive review of using deep learning for machine health monitoring is presented in [2]. However, the individual model sometimes has low generalization capability [4]. Therefore, ensembles of deep learning methods (EDL) for condition monitoring have been recently employed. In [5], Ma and Chu proposed an EDL for fault diagnosis. They made the ensemble by using different deep learning methodologies and then combined them by using multi-objective optimization. Chakraborty et. al. used stacked autoencoders to extract features and then use probabilistic neural networks to create an ensemble[6]. A majority voting method has been used as the combination method. In [4], Li et. al. created an ensemble model by using different variations of deep autoencoders and a weighted voting method. Zhang et. al. proposed to use non-negative sparse constrained deep neural networks as the base model for fault diagnosis [7]. The models are then combined by using the Dempster-Shafer theory of evidence.

 To employ an ensemble of deep learning (EDL) models, two questions should be addressed: i) how to impose diversity to the models in the ensemble, and ii) how to combine their outputs to make a final decision.

The first issue can be addressed by using different classification methods, different types of features, different training samples, etc. In this regard, two widely used approaches are bootstrap aggregation [8], and adaptive boosting [9]. However, the common practice is to generate a pool of classifiers and select some of them[10,11].

To tackle the second question, various methods have been developed. From basic operations to more advanced forms like Dempster-Shafer theory (DST) of evidence. Although DST has proven advantages over other combination methods, its main challenge occurs when conflicting evidences come from different models which could lead to counter-intuitive results. To treat this issue, one could employ different preprocessing techniques on the evidences. This could be done by employing confusion matrix [12], Shanon's information entropy [13], belief entropy and Fuzzy preference relations [14] at different stages of data preprocessing.

This study proposes an ensemble deep learning framework for vibration-based condition monitoring of complex metal parts. The framework is based on Convolutional Neural Network (CNN) and Dempster-Shafer theory (DST) of evidence.

## 2. Methodology

The method consists of four main steps, Data collection, CNN generation, CNN selection, and CNN combination. They are elaborated in the following.

### 2.1 Data collection

Data collection has been done with one actuator and two sensors. By that, the vibrational response data from Equiax Polycrystalline Nickel alloy first-stage turbine blades with complex geometry has been collected. Figure 1 shows two views of its CAD model. The cooling channel is visible in the transparent view. The amplitude of the frequency response function (|FRF|) has been collected from each blade in the range of [3, 38] kHz at 11253 frequency bins, see Figure 2. To create a database 150 healthy and 79 defected blades have been measured.

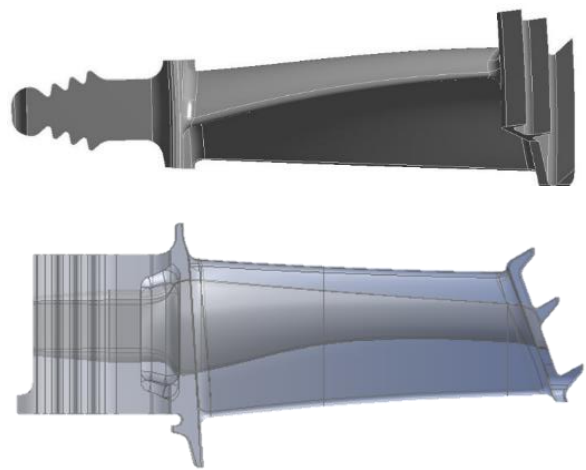

Figure 1. CAD model of the turbine blade. The cooling channel is visible in the transparent view.

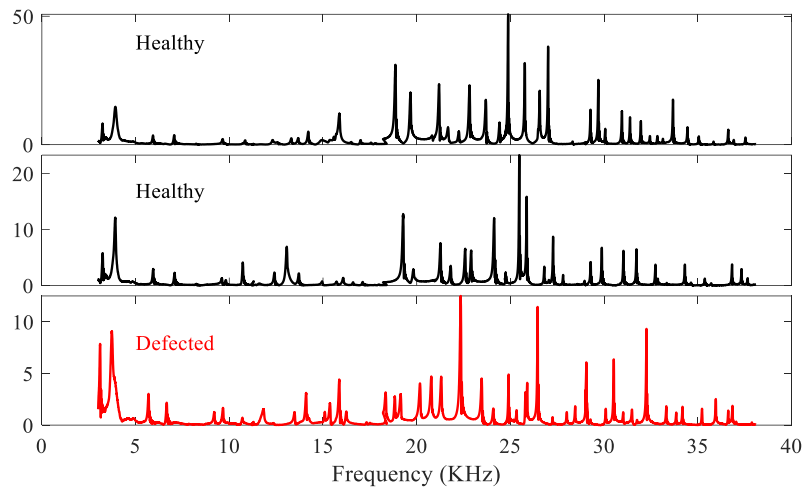

Figure 2. Three examples of the collected FRFs. Two top FRFs have been collected from healthy blades and the bottom one was from a defected blade.

*2.2   CNN generation*

In this section, a pool of CNN will be generated. For this purpose, one CNN structure with several different input signals has been used. The new CNN structure used here is shown in Figure 3. It consists of 16 layers and a softmax layer. The configuration of the layer is presented in **Table 1**.

To generate the pool, the frequency range of the FRFs is split into eight equal sections, each of which is used to train a CNN. Since each sample has 2 FRFs, this means, $2 \times 8 = 16$ CNNs are generated. Moreover, for each section of the FRF, another CNN is generated by using both signals. Therefore, in total $16 + 8 = 24$ CNNs are trained to make the pool.

The model is implemented using the Matlab 2019a, deep learning toolbox. For training, the adam optimizer default parameters are utilized. The batch-size was 128, the maximum used epochs set to 200, and the initial learning rate was 0.005 which decayed by the factor of 0.005 every 10 epochs.

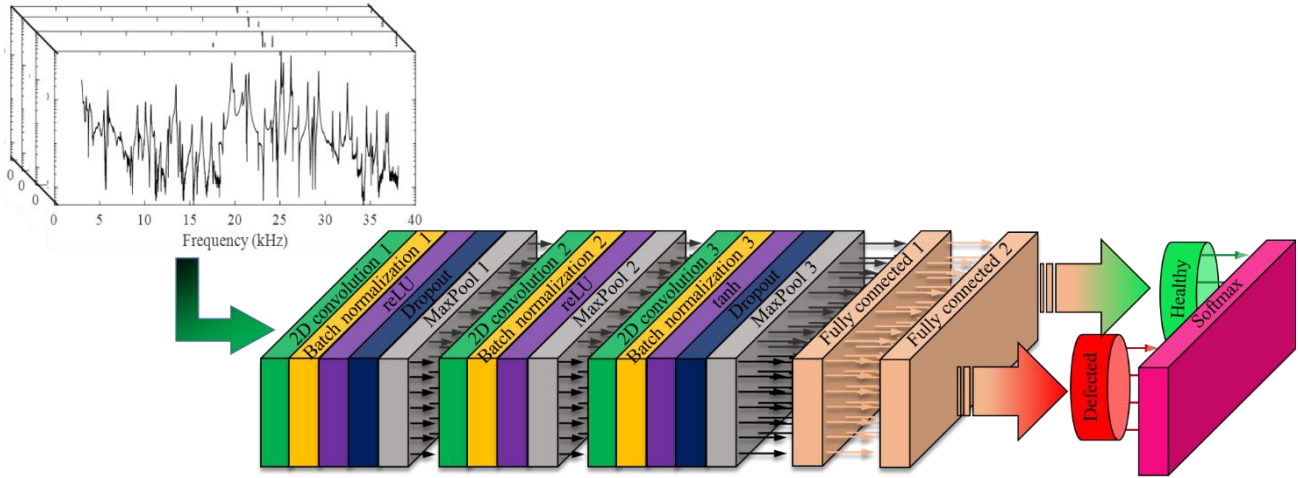

Figure 3. The proposed CNN structure

**Table 1. configuration of the layers in the proposed CNN**

| Layer | Kernel Shape | Kernel # | Stride |
|---|---|---|---|
| Convolution 1 | $1 \times 3$ | 64 | 1 |
| MaxPool 1 | 2 | — | 1 |
| Convolution 2 | $1 \times 3$ | 32 | 1 |
| MaxPool 2 | 2 | — | 1 |
| Convolution 3 | $1 \times 3$ | 16 | 1 |
| MaxPool 3 | 2 | — | 1 |
| FC 1 | 150 | 100 | — |
| FC2 | 100 | 2 | — |

*2.3 CNN selection/ranking*

Classifier ranking/selection is an inevitable step for creating a suitable classifier ensemble. In this work, it is done by using Mutual information between the classifiers output and the target outputs [15]. Let $\Gamma = (\gamma_1, \gamma_2, \ldots, \gamma_N)$, $\Omega = (\omega_1, \omega_2, \ldots, \omega_N)$ be two discrete random variables. Entropy $H$ measures the information available in each of these variables. It is defined as

$$H(\mathbf{\Gamma}) = -\Sigma_{i=1}^{N} p(\gamma_i) \log(p(\gamma_i)) \tag{1}$$

with $p(\gamma_i)$ is the probability mass function. Mutual information $(MI)$ is the amount of information between two random variables. It is defined as,

$$MI(\Gamma; \Lambda) = MI(\Lambda; \Gamma) = H(\Gamma) - H(\Gamma|\Lambda) \quad (2)$$

Here, $H(\Gamma|\Lambda)$ is conditional entropy indicating the amount of information in $\Gamma$ after introducing $\Lambda$. It is defined as,

$$H(\Gamma|\Lambda) = -\Sigma_{j=1}^{M}\Sigma_{i=1}^{N} p(\gamma_i, \lambda_j) \log(p(\gamma_j|\lambda_i)) \quad (3)$$

Now, let $Y \in \mathbb{R}^{n_s \times n_c}$ be the matrix of output in which, $n_s$, and $n_c$ are the number of samples, and classes. Further, let $\widehat{Y}_i, i = 1,2,...,N_c$ be the output predicted by the $i^{th}$ classifier. Then $MI(\widehat{Y}_i; Y)$ is used to rank the classifiers. One could set the number of models in the ensemble or monitor the validation accuracy of the ensemble to select the most accurate one. This is called the best ensemble model (BEM).

*2.4  CNN fusion by improved Dempster-Shafer theory*

In this section, the output of the CNNs in the ensemble will be combined to improve the classification performance. For this purpose, an improved Dempster-Shafer method is used [16].
Let $\mathfrak{E} = \{E_1, E_2, ..., E_K\}$ be a finite set of mutually exclusive and collectively exhaustive events. it is called the frame of discernment (FOD), $2^{\mathfrak{E}}$ is its power set, and $\mathcal{A}$ is called proposition provided that $\mathcal{A} \in 2^{\mathfrak{E}}$. Basic Belief Assignment (BBA) is a function that maps $\mathcal{A} \subseteq \mathfrak{E}$ to [0, 1] provided that

$$\begin{cases} m(\emptyset) = 0 \\ \sum_{\mathcal{A} \in 2^{\mathfrak{E}}} m(\mathcal{A}) = 1 \end{cases} \quad (4)$$

If $\mathcal{A} \subseteq \mathfrak{E}$ and $m(\mathcal{A}) \neq 0$, then $\mathcal{A}$ is called a focal element. For two independent BBAs $m_1$ and $m_2$ in the FOD $\mathfrak{E}$, the Dempster's rule of combination is defined as

$$m(\mathcal{A}) = (m_1 \oplus m_2)(\mathcal{A}) = \begin{cases} \frac{1}{1-K} \sum_{\mathcal{C}_1 \cap \mathcal{C}_2 = \mathcal{A}} m_1(\mathcal{C}_1) m_2(\mathcal{C}_2) & \mathcal{A} \subset \mathfrak{E}, \mathcal{A} \neq \emptyset \\ 0 & \mathcal{A} = \emptyset \end{cases} \quad (5)$$

$K$ is a measure of conflict between the two BOEs. It is defined as,

$$K = \sum_{\mathcal{C}_1 \cap \mathcal{C}_2 = \emptyset} m_1(\mathcal{C}_1) m_2(\mathcal{C}_2), \quad (6)$$

The combination rule can be applied to two BBAs if $K < 1$, otherwise the combination may give counterintuitive results. In such cases, the following preprocessing technique is used.
Step 1. Average belief divergence

$$aBJS_i = \frac{1}{N_c - 1} \sum_{\substack{j=1, \\ j \neq k}}^{N_c} \left[ \Sigma_i m_j(A_i) \log\left(\frac{2m_j(A_i)}{m_j(A_i) + m_k(A_i)}\right) + m_k(A_i) \log\left(\frac{2m_k(A_i)}{m_j(A_i) + m_k(A_i)}\right) \right] \quad (7)$$

Step 2. Evaluate the Support degree as

$$SD_i = \left(aBJS_i \times \left[0.5 + \frac{1}{\pi}\arctan\left(\frac{SW - SW_{\sim i}}{2}\right)\right]\right)^{-1} \quad (8)$$

in which

$$SW = \frac{1}{N_c}\Sigma_{j=1}^{N_c}\sqrt{\frac{|m_j|}{|\overline{m}|}(m_j - \overline{m})^T(m_j - \overline{m})} \quad (9)$$

$$SW_{\sim i} = \frac{1}{N_c - 1}\Sigma_{\substack{j=1\\j\neq i}}^{N_c}\sqrt{\frac{|m_j|}{|\overline{m}|}(m_j - \overline{m}_{\sim i})^T(m_j - \overline{m}_{\sim i})} \quad (10)$$

with $\overline{m}$ is the average of the BBAs, and $\overline{m}_{\sim i}$ is the average of the BBAs except for the $m_i$.

Step 3. Evaluate the credibility degree as

$$CD_i = \exp\left(-\Sigma_j m_i(A_j)\log(\frac{m_i(A_j)}{2^{|A_j|}-1})\right) \times \left[\frac{SD_i}{\Sigma_{j=1}^{N_c} SD_j}\right] \quad (11)$$

Step 4. Evaluate the weighted average evidences as

$$\boldsymbol{WAE} = \sum_{i=1}^{N_c}\left[\frac{CD_i}{\Sigma_{j=1}^{N_c} CD_j}\right] \times \boldsymbol{m_i} \quad (12)$$

Step 5. Combination based on Dempster's rule Eq. (5)

$$\widetilde{Y} = \bigoplus_{i=1}^{N_c} \boldsymbol{WAE} \quad (13)$$

## 3. Results and discussion

In this section, the ensemble CNN method has been applied to the |FRF| data. The dataset has been randomly split to [70, 30]% as the training and validation dataset. FRFs at the first sensor $X_1$, the second sensor $X_2$, and both sensors $(X_1, X_2)$ have been used as the input signal for training the 24 individual CNNs. Their validation accuracies are shown in **Table 2**. In this table, the first row shows the first frequency in the frequency range used for training the CNNs. The most accurate individual model with the accuracy of 94.20%, shown in bold, is obtained by using the FRF at the second sensor in the frequency range of [8.81 14.59] kHz. It is called the Best Individual Model (BIM). Moreover, the ranks of different CNNs for adding to the ensemble have been assessed and shown in the parentheses.

As the next step, the first model has been selected and the other models according to their rankings have been added to the ensemble. By each addition, the validation accuracy of the ensemble has been assessed and shown in Figure 4. The result indicates that the ensembles with three to six models with the accuracy of 95.65% have the best performance. They are called the best ensemble models (BEMs) and are shown in red in the figure. Moreover, it can be observed that adding inappropriate models to the ensemble could deteriorate the validation accuracy of the ensemble.

Table 2. Validation accuracy of the individual CNNs trained with the FRF at the first sensor $X_1$, second sensor $X_2$, and both sensor $(X_1, X_2)$. The first row indicate the frequency range used for training in kHz. The number in the parentheses is the rank of each classifier.

| Freq. (kHz) | 3.00 | 8.81 | 14.59 | 20.46 | 25.70 | 28.92 | 32.12 | 35.41 |
|---|---|---|---|---|---|---|---|---|
| $X_1$ | 91.30 (4) | 92.75 (2) | 91.30 (7) | 91.30 (8) | 81.16 (16) | 79.71 (17) | 86.96 (13) | 65.21 (24) |
| $X_2$ | 91.30 (6) | **94.20 (1)** | 88.41 (11) | 89.86 (9) | 73.91 (22) | 68.11 (21) | 88.40 (10) | 73.91 (20) |
| $(X_1, X_2)$ | 92.75 (3) | 91.30 (5) | 86.95 (4) | 86.96 (12) | 76.81 (19) | 79.71 (18) | 84.06 (15) | 65.21 (23) |

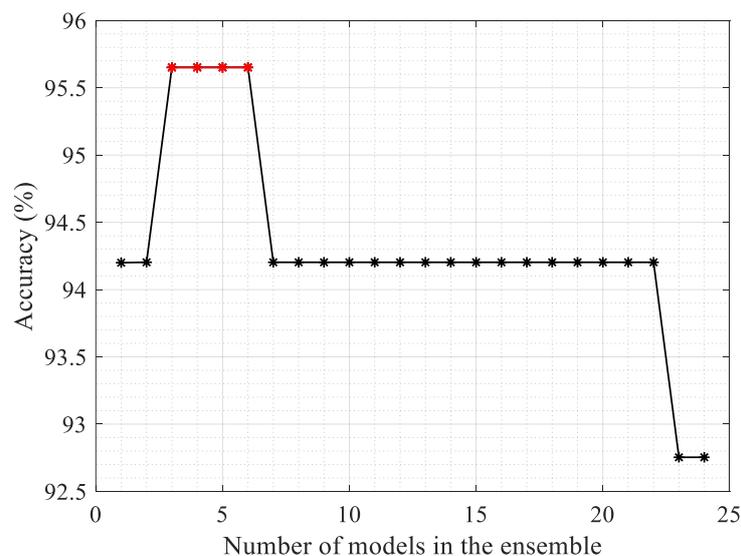

Figure 4. Validation accuracy of the ensemble CNN. The best ensemble models are shown in red.

Table 3. Validation accuracy of the CNNs trained with the full FRFs and their associated BEM. The ranking of the individual CNNs is in parentheses.

| FRF | $X_1$ | $X_2$ | $(X_1, X_2)$ | BEM |
|---|---|---|---|---|
| Accuracy | 92.75 (2) | 91.30 (3) | **94.20 (1)** | 94.20 |

A comparison has been made between the performance of the proposed ensemble framework with the case that the full FRFs used for training the CNNs. The validation accuracy of the CNNs trained on the full FRFs is shown in **Table 3**. One can see that in this case, the CNN trained on both full FRFs gives the highest accuracy which is equal to the accuracy of the BIM in **Table 2**. Besides that, the CNNs has been ranked and combined based on the presented approach. Their associated BEM gives the accuracy of 94.2%. This comparison implies that by employing the presented ensemble framework, not only do we lose any accuracy in the individual models, but we also impose diversity to the ensemble leading to improving the classification performance of the BEM.

In the end, to perform statistical analysis, the whole framework has been performed 20 times. At each repetition, the CNN with the highest validation accuracy, i.e. BIM, has been obtained. Then the CNNs have been combined to generate the ensemble with the highest validation accuracy, i.e. BEM. The result of each repetition is shown in Figure 5. As can be seen, at all repetitions but the 3$^{rd}$ one, the validation accuracy has improved. This thus leads to improving the average classification performance from 94.05 for the BIMs to 96.28% for the BEMs.

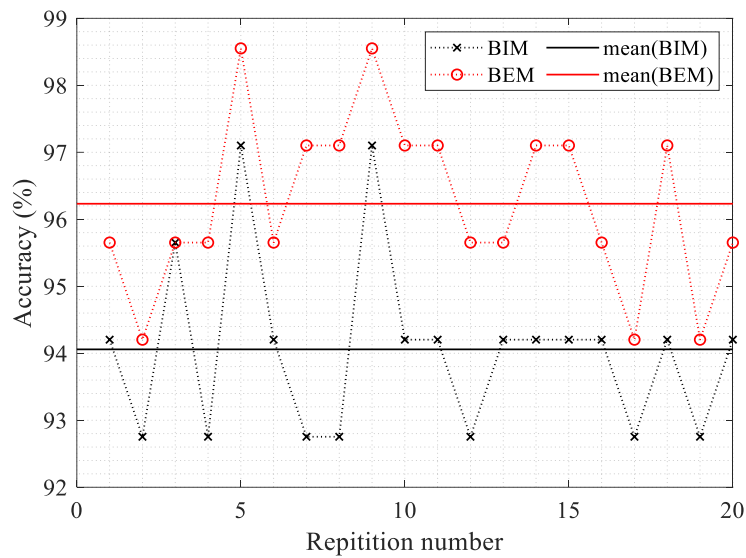

Figure 5. Accuracy comparison between the Best individual models (BIMs) and their associated Best ensemble models (BEMs) through 20 repititions of the ensemble framework. The solid lines indicate the mean-values of the accuracies.

## 4. Conclusion

In this paper, an ensemble deep learning framework has been presented to be used for vibration-based condition monitoring. It has been achieved by using a convolutional neural network (CNN) in conjunction with the Dempster-Shafer theory of evidence (DST). The framework consists of four stages: (i) Data collection in the form of FRF. (ii) CNN generation to make a pool. This has been obtained by splitting the FRF into several sections, each of which has been used to train a CNN. (iii) CNN selection from the pool by using the mutual information criterion. (iv) CNN combination by using an improved DST. The framework has been applied to a vibration dataset collected from Nickel-alloy first-stage turbine blades with complex geometry. The results indicated that the presented ensemble framework could improve the average classification performance from 94.05%, for the best individual CNN, to 96.28%.

## 5. Acknowledgment
The authors gratefully acknowledge the ICON project DETECT-ION (HBC.2017.0603) which fits in the SIM research program MacroModelMat (M3) coordinated by Siemens (Siemens Digital Industries Software, Belgium) and funded by SIM (Strategic Initiative Materials in Flanders) and VLAIO (Flemish government agency Flanders Innovation & Entrepreneurship). Vibrant Corporation is also gratefully acknowledged for providing anonymous datasets of the blades.